\documentclass[smallextended]{svjour3}       % onecolumn (second format)

\smartqed  % flush right qed marks, e.g. at end of proof
\usepackage{graphicx}
\usepackage{graphicx}
\usepackage{times}
\usepackage{helvet}
\usepackage{courier}
\usepackage{amsmath}
\usepackage{algorithm}
\usepackage{algorithmic}
\usepackage{csquotes} 
\usepackage{color}
\usepackage{paralist}
\usepackage{amssymb}
\usepackage{indentfirst}
\usepackage{subfigure}
\usepackage{float}
\usepackage{multirow}
\usepackage{cite}
\usepackage{hyperref}
\usepackage{mathrsfs}

\begin{document}

\title{ Improved Hard Example Mining by Discovering Attribute-based Hard Person Identity }

%\titlerunning{Short form of title}        % if too long for running head 

\author{Xiao Wang \and Ziliang Chen \and Rui Yang \and Jin Tang \and Bin Luo}

%\authorrunning{Short form of author list} % if too long for running head

\institute{ Xiao Wang, Rui Yang, Bin Luo, Jin Tang \at
              School of Computer Science and Technology, Anhui University, 111 Jiulong Road, Shushan District, Hefei, Anhui, China \\
              \email{\{wangxiaocvpr, yangruiahu\}@foxmail.com, \{luobin, tangjin\}@ahu.edu.cn}           %  \\
           \and
           Ziliang Chen \at 
           School of Data and Computer Science, Sun Yat-Sen University, Guangzhou, China. \\
           \email{c.ziliang@yahoo.com}
}

\date{Received: date / Accepted: date}
% The correct dates will be entered by the editor

\maketitle

\begin{abstract}
Hard example mining is a sophisticated technique widely applied in person re-IDentification (re-ID): Given an image of one person, the hard example mining searches its closet neighbour sample that belongs to the other person, then forms the dissimilarity-based embedding couples to train the deep models with triplet metric losses. But the current hard negative examples are limitedly searched in a mini-batch, where only a few of person identities are locally included. As the population increases, this routine becomes less efficient to find out the global hard example across all training identities for constructing each triplet loss. In this paper, we propose Hard Person Identity Mining (HPIM) that attempts to refine the hard example mining to improve the exploration efficacy. It is motivated by following observation: the more attributes some people share, the more difficult to separate their identities. Based on this observation, we develop HPIM via a transferred attribute describer, a deep multi-attribute classifier trained from the source noisy person attribute datasets, by semi-supervised learning following the attribute grouping manner. We encode each image into the attribute probabilistic description in the target person re-ID dataset. Afterwards in the attribute code space, we consider each person as a distribution to generate his view-specific attribute codes in different practical scenarios. Hence we estimate the person-specific statistical moments from zeroth to higher order, which are further used to calculate the central moment discrepancies between persons. Such discrepancy is a ground to choose hard identity to organize proper mini-batches, without concerning the person representation changing in metric learning. It presents as a complementary tool of hard example mining, which helps to explore the global instead of the local hard example constraint in the mini-batch built by randomly sampled identities. We validated the method on two popular person re-ID benchmarks CUHK-03 \cite{li2014deepreid} and Market-1501 \cite{zheng2015scalable}, which both demonstrate the efficacy of our model-agnostic approach. We use PETA \cite{deng2014pedestrian} and extra unlabelled noisy data sources to attain our attribute encoder, which also outperforms various existing baselines in attribute recognition.

\end{abstract}

%	==============================================================
%	=============		Introduction	       ===========================
%	==============================================================
\section{Introduction}

\begin{figure}[t]
\centering
\includegraphics[width=3.5in]{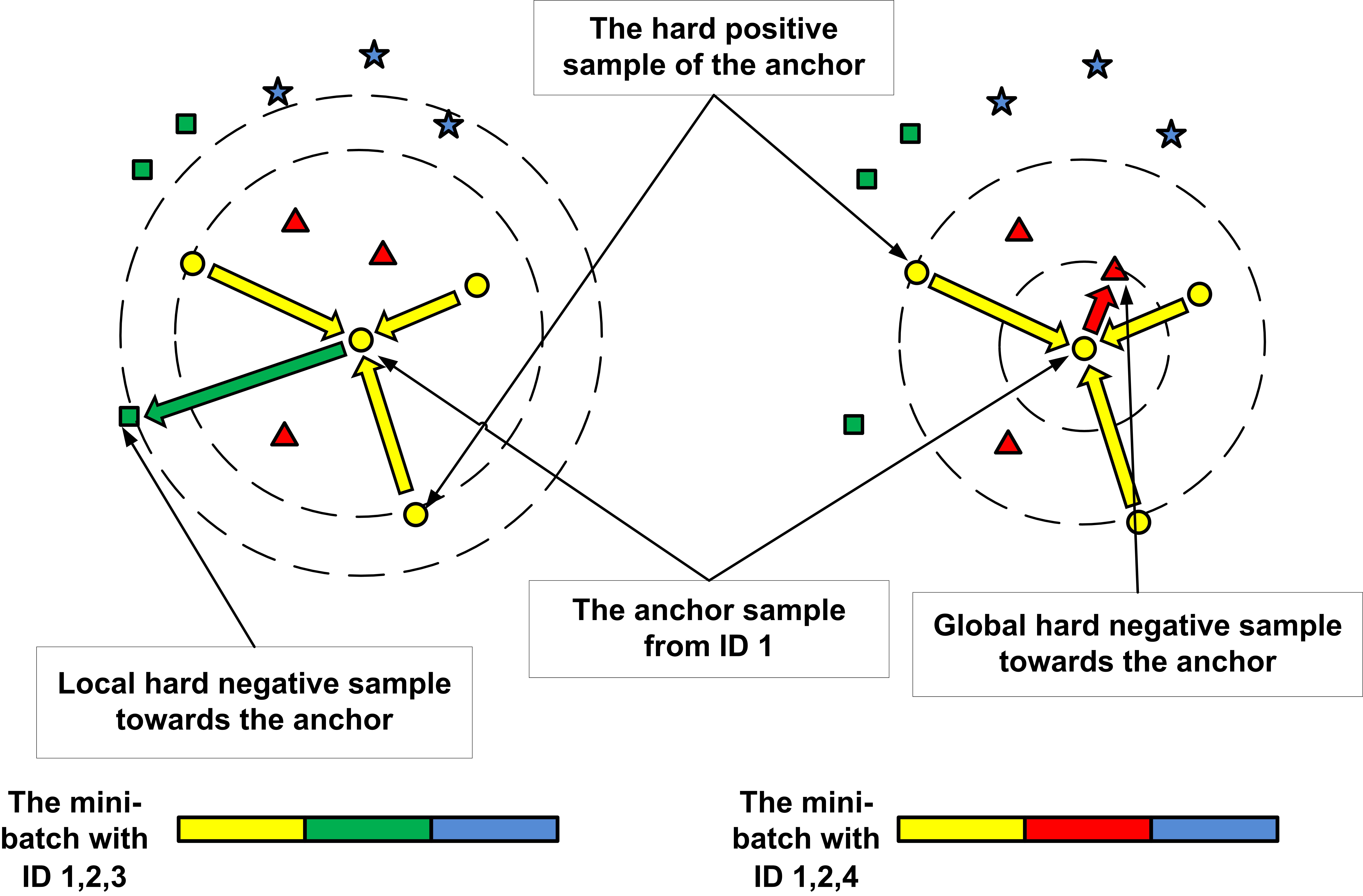}
\caption{ The illustration of traditional local hard example mining and our proposed global hard negative mining mechanism. }
\label{frontImage}
\end{figure}	

%% ===== what, how and why doing the Re-ID and corresponding popular algorithms.
Person re-identification aims to match pedestrians observed from non-overlapping camera views based on the visual appearance. The problem helps to search for a person from large amounts of images and videos easier, hence, attracts enormous attentions due to its wide range of potential applications, e.g., human retrieval, tracking and activity analysis. However, since remotely captured visual data usually suffer from blurring, background clutters and occlusions, the problem remains quite challenging in practice. Recently deep representation learning and metric learning present prevalent for solving Re-ID. The former discovers the visual matching information without hand-crafted manipulation. The latter opts for optimizing the semantic embedding space so that the data points with the same identity keep closer than the data points with different identities. Both of them naturally suit the setting of person re-ID and their cooperation, deep metric learnings, have become the modern go-to solutions \cite{Zhu2017Deep} \cite{vezzani2013people} \cite{bedagkar2014survey}.

%% =====	hard example mining ===== from deep metric learning to hard example mining.
In most literatures, deep metric learning performs stochastic to optimize the embedding space. The procedure necessitates semantic-meaningful embedding combinations to reduce the sequential learning variance. Hard example mining algorithm is broadly accepted to response to this concern \cite{cheng2016person} \cite{chen2015similarity} \cite{ding2015deep}. Its key idea derives from the variants of the triplet loss (a sort of embedding optimization manner). Specifically, the algorithm firstly pre-processes training batches by randomly selecting some people's identities so as to construct each batch that incorporates the images with the corresponding identities. Then given an image as the \emph{anchor}, the algorithm chooses a couple of the other images considered within the same batch, with the same identity while the embedding located farthest (hard positive) and with the different identity (hard negative) while the embedding located nearest, respectively. Each anchor with its hard positive and negative examples organizes a triplet loss to facilitate the directed stochastic training. Its successes have already witnessed in an extensive array of practices.

%% ==== the shortcoming of hard example mining ====			
But as we previously mentioned, since training batches are pre-processed by the random selection of identities, the hard negative example discovery only involves a few of people, instead of a consideration across the total training population. It probably causes \textbf{a risk of selecting the identities presenting locally hard yet globally easy for optimizing the embedding space}. As illustrated in Fig.1 (left), the embeddings with the hard identity (green) within a batch obviously show distinct from the embeddings with the identity (yellow) the anchor belongs to. Concurrently, the global hard identity (red) contains the examples with more confusing embeddings yet has been ironically ignored. This situation harms the efficiency of hard example mining and as the identity number grows or the batch size decreases, becomes dominant during the triplet-based stochastic metric learning.

To overcome the deficiency, we develop Hard Person Identity Mining (HPIM) algorithm to replace the random identity selection in hard example mining. It treats people's identities as attribute-based embedding-invariant distributions, which are employed to calculate their statistical moment discrepancies as the criterion to select hard person identity. HPIM derives from a straightforward observation: people with more visual realism (person attributes) in common should be more difficult to be told apart. Such attribute-based person identity discovery perceives a merit that, we are able to choose hard person identities regardless of the varying embeddings, which makes the identity relationships prohibitively calculated in the large-scale data setting. For this purpose, our improved hard example mining includes two modifications: \textbf{1)}. Accept an classifier to produce the attribute description in probability for each image, which are robustly trained with penny-a-line attribute annotated datasets, namely, should be tolerant with massive and low-quality attribute annotations; \textbf{2)}. Treat each person as a distribution of his image-based attribute descriptions, thus, calculate the high-order statistical moments used to compute the \emph{central moment discrepancies} (CMD) between different people. Then given an anchor identity of someone, the other with less CMD value is the corresponding harder identity since it means the latter person look more similar with the anchor person. Hence, the latter identity is more probably chosen to organize a mini batch with this anchor identity. 
	
More specifically, we leverage a transferred human attribute describer to produce the visual description code of each image in the re-ID dataset. The describer is a multi-label classifier trained in another person-related datasets labelled with visual attributes. Meanwhile, we adopt the unlabelled human images to obtain more accurate attributes classifier in a semi-supervised manner. In another word, we optimize the attribute encoding network with training samples from easy to hard. After the attribute probabilities are predicted by the classifier, we are able to obtain the probabilistic attribute description for each images. We assume each person conceives a latent identity distribution to generate its attribute code. Think reversely and we plan to measure the hardness among persons by measuring their latent distribution distance in statistic. Concretely, we obtain the person-specific statistical moments from zeroth to higher orders, then leverage them to calculate the central moment discrepancies among person distributions. The discrepancies are used to choose hard identity to assemble each mini-batch, without concerning the changing visual embeddings in metric learning. It presents an complementary tool of hard example mining, which helps to explore the global instead of the local hard example constraint in the mini-batch built by randomly sampled identities.
	
%{\color{red} HPIM shows the merits to efficiently select hard identities. Firstly, the attribute codes keep consistent with the image instead of the embedding. It means for each time we assemble the batch, we don't have to go through the embedding space to recalculate the . }

The contributions of this paper can be concluded as the following three aspects:

\begin{itemize}
\item We propose an efficient hard person identity mining algorithm via the model-based attribute transfer for person re-identification. Specifically, global hard example mining is introduced to construct more efficient mini-batches and a new criterion is proposed to further improve the re-ID learning performance.

\item We propose a group-wise semi-supervised multi-label learning mechanism for human attribute recognition. We intuitively take the correlations between person attributes (\emph{i.e.} mutually exclusive and complementary) into the estimation pipeline when dealing with noisy unlabelled samples. And the learning procedure start from preliminary pre-trained models based on few labelled samples, and gradually mining unlabelled samples into the training set in an easy to complex way. 

\item Extensive experiments on both attributes recognition (PETA \cite{deng2014pedestrian}) and person Re-ID benmarks (Market1501 \cite{zheng2015scalable} and CUHK03 \cite{li2014deepreid}) all validate the effectiveness of our proposed methods. 

\end{itemize}

% =============================== 		 Related Work
\section{Related Works}\label{RelatedWorks}
In this section, we will give a brief review on person re-identification, hard example mining and pedestrian attributes learning. 

\textbf{Person Re-identification.~}
Appearance modelling and metric learning are the two key points in person re-identification to establish correspondences between person images. Many features are designed for person re-identification. An ensemble of local features (ELF) \cite{Gray2008Viewpoint} is constructed by using the eight color channels corresponding to the three separate channels of the RGB, YCbCr and HSV color spaces with the exception of the value channel. Other method is the utilization of local maximal occurance feature (LOMO) based on multi-scale Retinex to estimate HSV color histograms used for color features \cite{Liao2015Person}. The scale invariant local ternary pattern (SILTP) descriptor is used to model illumination invariant texture \cite{Liao2010Modeling}. Color and texture features are usually concatenated to form a high dimensional feature vector which is used as an input for learning methods. Deep features also has been validated its powerful representation ability in many tasks \cite{he2016deep} \cite{liu2015semantic} \cite{girshick2015fast} \cite{tao2016siamese} \cite{lecun2015deep}, and state-of-the-art person re-ID algorithms nearly all adopt the deep features.
	
However, only appearance comparison may failed in some challenging cases, hence, attributes based Re-ID algorithms attempt to handle this problem by incorporating semantic attributes. Pedestrian attributes (such as \emph{'female'}, \emph{'age'}, \emph{'hair style'} and \emph{'carry objects'}) can improve the performance of person Re-ID significantly. These attributes can be seen as mid-level features, and are usually learned from a large dataset with annotated attributes \cite{deng2014pedestrian}. Li \emph{et al.} present a comprehensive study on clothing attributes assisted person re-identification in \cite{li2015clothing}. They adopt latent SVM to describe the relations among the low-level part features, middle-level clothing attributes, and high-level re-identification labels of person pairs for robust person re-identification.

Metric learning aims to minimize the distances between the images drawn from the same classes while maximize the distances between the images drawn from different classes \cite{Jurie2013PCCA} \cite{Hirzer2012Relaxed} \cite{Zheng2016MARS}. The KISSME \cite{Hirzer2012Large} and XQDA\cite{Liao2015Person} are two widely used metric learning algorithms for person Re-ID based on Mahalanobis distance. KISSME casts the problem in the space of pairwise differences on a likelihood ratio test. XQDA extends Bayesian faces and KISSME by learning a subspace reduction matrix and a cross-view metric through a joint optimization. The closed-form solution is obtained by formatting the problem as a generalized Rayleigh quotient and using eigenvalue decomposition \cite{Liao2015Person}. Recently, deep neural networks (DNNs) are introduced into the person re-ID community due to the release of large datasets and also widely applied in many other tasks \cite{wang2019quality, wang2019GANTrack, zhu2019denseTracking, zhai2019fmt}. The popular network architecture search technique is also used to design more powerful network for person re-ID \cite{quan2019autoreid}. The overall performance of person Re-ID is promoted to a higher level.

\textbf{Hard Example Mining} is a commonly used technique to improve shallow re-ID models, which usually starts with a dataset of positive examples and a random set of negative examples. A model is then trained on that dataset until convergence and subsequently executed on a larger dataset to obtain false positive samples. The training set will absorb these samples and will be used to re-train the machine learning model again. Some recent works \cite{Simo2014Fracking} \cite{Xiaolong2015Unsupervised} \cite{Loshchilov2015Online} also focus on selecting hard examples for training deep networks as our proposed algorithm. These methods select their samples according to current loss for each data point. Specifically, given a positive pair of patches, Wang \emph{et al.} select hard negative patches from a large set using triplet loss in \cite{Xiaolong2015Unsupervised}. Loshchilov \emph{et al.} investigates online selection of hard examples for mini-batch SGD methods in \cite{Loshchilov2015Online}.  Florian \emph{et al.} use triplets of roughly aligned matching/non-matching face patches generated using a novel online triplet mining method in \cite{Schroff2015FaceNet}. Many works has been extended based on the triplet loss function in person re-identification, such as \cite{Hermans2017In} \cite{Chen2017Beyond}. 
	
Some researches also generate hard examples to enhance the learning across the other tasks \cite{wang2017fast} \cite{Wang_2018_CVPR} \cite{Zhong_2018_CVPR} \cite{Deng_2018_CVPR}. Wang \emph{et al.} \cite{wang2017fast} propose a novel method to help improving robustness of deep neural networks via an adversarial network that generates occluded and deformed examples. Wang \emph{et al.} \cite{Wang_2018_CVPR} propose to generate hard positive samples via adversarial learning for visual tracking due to the sparse of hard positive samples in practical training dataset and dense sampling strategy. Deng \emph{et al.} \cite{Deng_2018_CVPR} propose to use domain adaptation policy to improve person re-ID with similarity preserving generative adversarial network (SPGAN) which consist of an Siamese network and a CycleGAN. Zhong \emph{et al.}  \cite{Zhong_2018_CVPR} introduced the similar style transfer techniques by explicitly address the image style variation issues in person re-ID which also can be seen as a kind of hard sample generation. These works all demonstrate the effectiveness of hard samples in practical applications.  However, seldom of them considered the hard mini-batch construction when training their network.

%% ===============================
\textbf{Pedestrian Attributes Learning.~}
Earlier person attribute recognition model multiple attributes independently and train a specific classifier for each attribute based on hand-crafted features such as colour and texture histograms \cite{Escorcia2015On} \cite{G2014Actions} \cite{Bourdev2009Poselets}.  Some graph models, such as CRF (conditional random field) or MRF (Markov random field), are utilized to model the inter-attribute correlation for improving recognition performance \cite{Chen2012Describing} \cite{Deng2015Learning} \cite{Shi2015Transferring}. But when dealing with a large set of attributes, these methods are computational expensive due to the huge number of model parameters on pairwise relations. Recent works has shown that joint multi-attribute feature and classifier learning \cite{Zhu2015Multi} \cite{Zhu2017Multi} \cite{sudowe2015personworkshop} \cite{Li2016Human} \cite{Li2016Multi} \cite{dong2017class} to benefit from learning attribute co-occurrence dependency. And other algorithms also exploited contextual information \cite{Gkioxari2015Contextual} \cite{Li2016Human} in their framework, however, making too strong assumptions about image qualities to be applicable to surveillance data. Wang \emph{et al.} \cite{Wang2017Attribute} introduce  a joint recurrent learning model for exploring attribute context and correlation in order to improve attribute recognition given small sized training data with poor quality images. Zhao \emph{et al.} propose the concept of ``attribute group" and integrate it with recurrent neural network in \cite{zhao2018grouping}.	More pedestrian attribute recognition algorithms can be found in \cite{wang2019pedestrian}. However, the performance of aforementioned works is still limited because their model is trained on limited annotated dataset only. Meanwhile, our person attribute recognition algorithm is pre-trained on little annotated dataset and then self-promoted based on large scale unlabelled images in a semi-supervised manner. Our algorithm also achieve comparable or even better performance than existing approaches.

%% ============================================= 
%% =============== The Proposed Approach ========= 
%% ============================================= 

\section{The Proposed Method}\label{theAllAlgorithm}
In this section, we will first revisit the setting of re-ID based on deep metric embedding learning. Then, we will give an overview of our improved hard example mining by hard identity mining which is a two-stage learning framework bridging robust attribute recognition and person re-ID. Afterwards, we detail the person attribute describer, which are robustly trained in a noisy environment by combining the attribute grouping and self-paced learning (SPL) techniques. We then discuss the hard identity discovery based on the attribute descriptions and central moment discrepancy (CMD), and present a CMD-based policy sampling approach to refine the hard example mining.

\subsection{\textbf{Re-ID with Metric Embedding Learnings}}
We begin with a vanilla Re-ID setup: There are $N$ person identity sets $\{X_1, \cdots, X_{N}\}$, where $X_i=\{x^{(i)}_s\}^S_{s=1}$ includes $S$ images of someone with the $i^{th}$ person's identity. Re-ID aims to find out a mapping $f_{\theta}$ w.r.t. the parameter $\theta$, therefore given each image $x^{(i)}_s$ that belongs to the $i^{th}$ person, $f_{\theta}(x^{(i)}_s)$ can refer to $f_{\theta}(x^{(i)}_{s'})$ ($\forall s'\in[S]$). It is emphasized that, the identities during training and testing phases are not overlapped, thus, visual recognition models hardly suit this problem.

Metric embedding learnings (MELs) are employed to solve person re-ID. Roughly speaking, suppose $f_{\theta}(x)$ denotes the embedding of image $x$ extracted from the model $f_{\theta}(\cdot)$ parameterized by $\theta$. Given a specific anchor image $x^i_{s_1}$ ($\forall i\in[N]$, $\forall p_1\in[S]$), MELs optimize the embedding space to minimize the distances between the samples with the same identities, e.g., $\ell(f_{\theta}(x^i_{s_1}),f_{\theta}(x^i_{s_2}))$ where $\forall s_2\in[S]/\{s_1\}$, and maximize the distances between the samples with different identities, e.g. $\ell(f_{\theta}(x^i_{s_1}),f_{\theta}(x^j_{s_3}))$ where $\forall s_3\in[S]/\{s_1\}$, $\forall j\in[N]/\{i\}$. There are plenty of MEL variants developed from this leitmotiv and we focus on the triplet-based MELs, a MEL family widely applied in most literatures. Given $P$ identities selected in each batch, the objective of triplet-base MEL can be formulated as: 
\begin{equation}\label{eq1.1}
\begin{small}
\mathcal{L}(\theta; X) = \sum_{i=1}^{P}\sum_{a=1}^{S}[m + \mu \underset{s=1}{\sum^{S}} \ell(f_{\theta}(x^i_a),f_{\theta}(x^i_s))  - (1-\mu) \underset{j=1\&j\neq i}{\sum^{P}}\sum_{s=1}^{S} \ell(f_{\theta}(x^i_a),f_{\theta}(x^j_n))]_{+}
\end{small}
\end{equation} 
where $m$ denotes the margin used to separate different identities and $\mu$ balances the strength between pushing and pulling operations for each anchor embedding. $\ell(\cdot,\cdot)$ measures the discrepancy of two embeddings:  
\begin{equation}\label{eul}
\ell(f_\theta(x_1), f_\theta(x_2)) = d(f_\theta(x_1), f_\theta(x_2)), \forall x_1,x_2\in \overset{N}{\underset{i=1}{\cup}} X_i
%\ell(x,y,\alpha) = \log(1+e^{\alpha d(x,y)})
\end{equation}
where $d(\cdot,\cdot)$ denotes the Euclidean distance. Eq.\ref{eq1.1} also evolves to the exponential variant called \emph{Lifted Embedding loss} as follows: 
    \begin{equation}\label{eq1.2}
    \begin{small}
    \mathcal{L}(\theta; X) = \sum_{i=1}^{P}\sum_{a=1}^{S}[ \mu \underset{s=1}{\sum^{S}} e^{\ell(f_{\theta}(x^i_a), f_{\theta}(x^i_s))} + (1-\mu) \underset{j=1\ j\neq i}{\sum^{P}}\sum_{s=1}^{S}e^{m- \ell(f_{\theta}(x^i_a), f_{\theta}(x^j_s))}]_{+}
    \end{small}
    \end{equation}
Eq.\ref{eq1.1} and \ref{eq1.2} make sure that the positive points with the same person identity project closer to the anchor's embedding than that of
a negative point with another identity, by $m$ magnitude at least. Thus, given an image in the testing gallery, we can discover its embedding neighbours as the images sharing the identity.

\textbf{Hard Example Mining for MEL.} The triplet-based MELs consider all anchor-positive and anchor-negative pairs in each batch during training. But in practice, the cubic triplet combination leads to a cumbersome and inefficient training process. Hard example mining is then employed to ease the problem. Concretely, it first randomly selects some people's identities to ensure each batch containing all the images corresponding to these identities. Hence given the embedding of an anchor image in each batch, the algorithm chooses a couple of the other images within this batch, with the same identity yet the embedding located farthest (hard positive) and with the different identity yet the embedding located nearest (hard negative), respectively. Under this principle, Eq. \ref{eq1.1} and Eq. \ref{eq1.2} can be reformulated as: 
    \begin{equation}\label{eq2.1}
    \begin{small}
    \mathcal{L}(\theta; X) = \sum_{i=1}^{P}\sum_{a=1}^{S}[m \ + \ \underset{s=1,\cdots,S}{\max} \ell(f_{\theta} (x^i_a),f_{\theta}(x^i_s)) - \underset{\substack{j=1,\cdots,P \\ s = 1,\cdots,S \\ j\neq i}}{\min}  \ell(f_{\theta}(x^i_a),f_{\theta}(x^j_s))]_{+}
    \end{small}
    \end{equation}
    and 
    \begin{equation}\label{eq2.2}
    \begin{small}
    \mathcal{L}(\theta; X) = \sum_{i=1}^{P}\sum_{a=1}^{S}[ \mu \underset{s=1,\cdots,S}{\max} e^{\ell(f_{\theta}(x^i_a),f_{\theta}(x^i_s))} + (1-\mu)\underset{\substack{j=1,\cdots,P \\ s = 1,\cdots,S \\ j\neq i}}{\min}e^{m- \ell(f_{\theta}(x^i_a),f_{\theta}(x^j_s))}]_{+} 
    \end{small}
    \end{equation}

\textbf{Intuition of HPIM.} In fact, the wisdom of hard example mining maintains an inconspicuous flaw, which might causes a serious problem as the identity number $N$ increases. Obviously in Eq. \ref{eq2.1} and \ref{eq2.2}, the hard examples are basically discovered within a batch, namely, the hard negative examples are merely extracted from $P$ instead of $N$ identities ($P<<N$). But as the training person identity number increases, $P$ identities collected in each batch can spread quite diverse. Even though the embedding of the hard negative example keeps locally (in a batch) close to the anchor, whereas, it still can be located far away from the anchor over the training embeddings with all identities. Since the embeddings of all examples keep varying as $f_{\theta}(\cdot)$ stochastically updates, it is prohibitive to estimate the relationships of identities by the moving statistic of those changing embeddings. This challenge obstacles the scalability and effectiveness of hard example mining.

\begin{figure*}[t]
\center
\includegraphics[width= 4.5in]{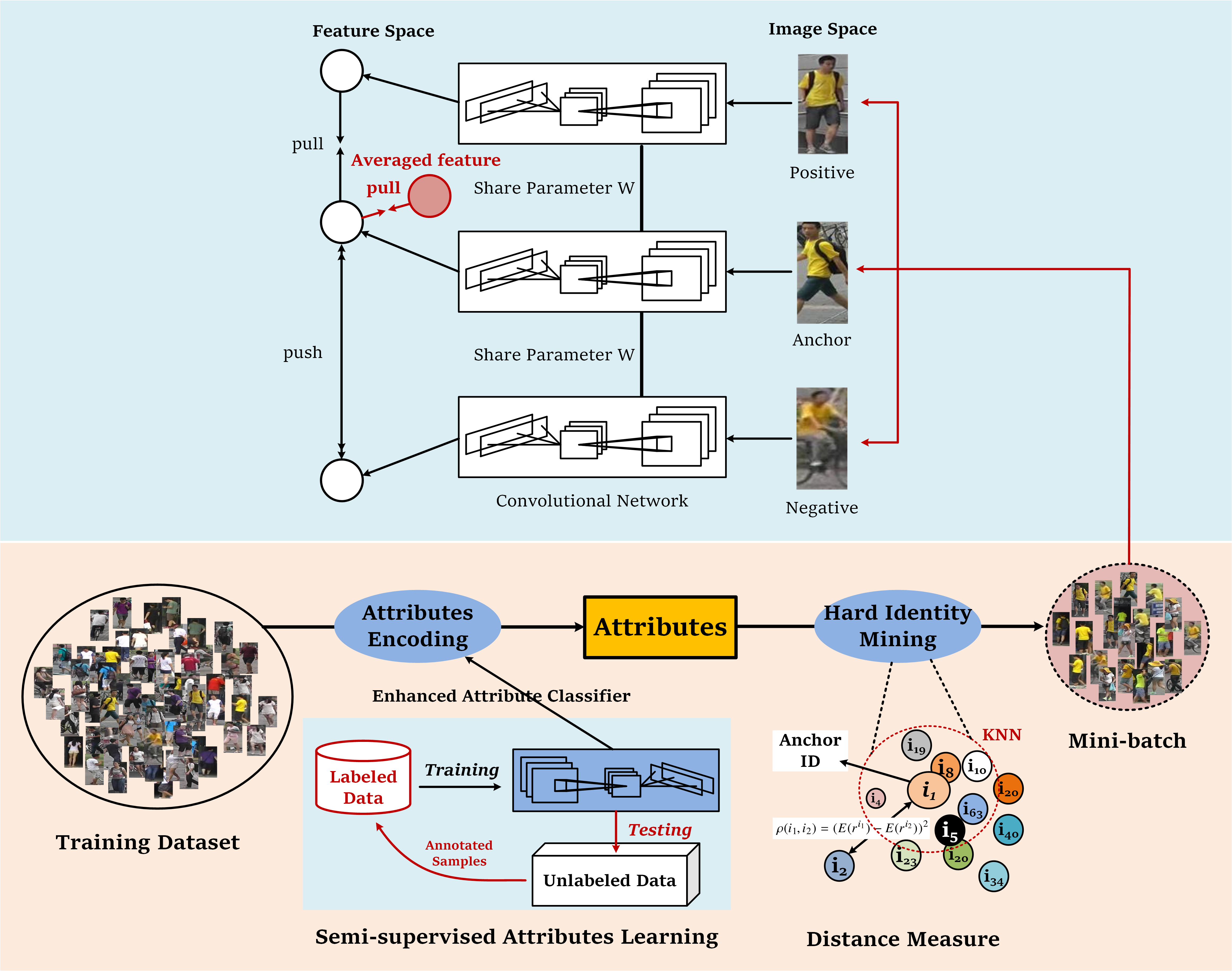}
\caption{The pipeline of our proposed person attribute assisted re-identification algorithm.}
\label{wholePipeline}
\end{figure*}

Our solution aims to construct each training batch, through \emph{purposefully} instead of \emph{randomly}, selecting identities. In order to avoid redundant computation about the relationships among identities, we tend to leverage the information coming from the people's visual descriptions rather than their varying semantic embeddings. Hence our improved hard example mining modifies the triplet-based MEL as a two-stage learning framework whose work flow can be found in Fig. \ref{wholePipeline}. In the stage one, we import the other person dataset $\overline{X}$ labelled by multiple attributes to train an attribute describer (a multi-label classifier) providing a visual description code for each image. In the stage two, we treat each person identity as a distribution over the description codes of the person-related images. Computing the statistical moments for each person, we are able to estimate the central moment discrepancies (CMDs) between each pair of identities during training. The CMD keeps independent with the changing embeddings, thus, can be reused to arrange the online identity collection batch by batch without any extra computation. We finally propose a bandit-based policy based on the CMD results, which stochastically samples hard identities to construct the training batches for Re-ID.

\subsection{\textbf{Stage One: Attribute Describer Encoding}}\label{GSPL}
\begin{figure}[t]
\centering
\includegraphics[width=4in]{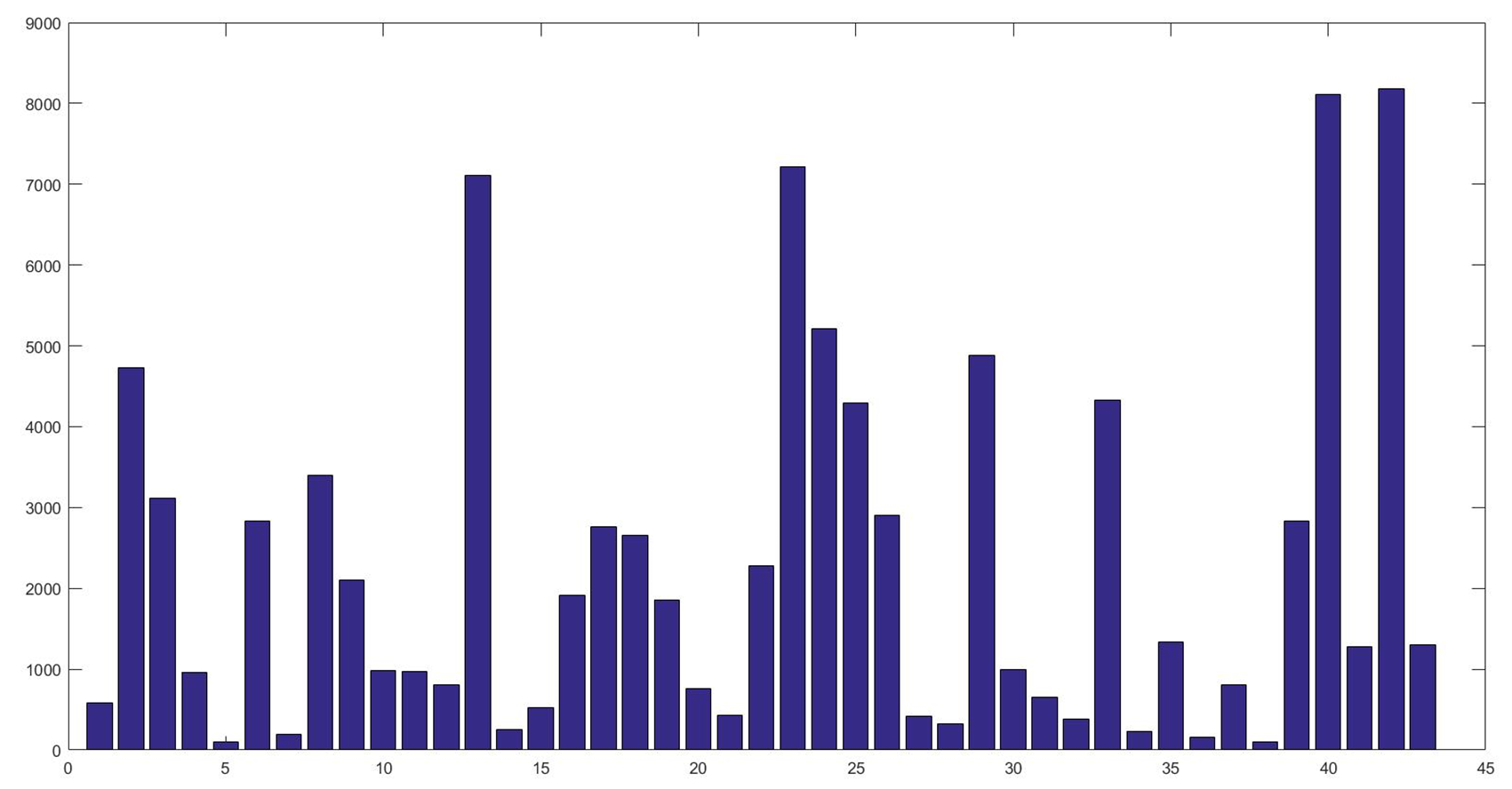}
\caption{ The unbalanced data distribution of different person attributes. }
\label{dataDistribution}
\end{figure}

In order to incorporate embedding-agnostic visual realism of each image, we need denser attribute annotations for providing more visual realism among people in details. But the dilemma is, denser attribute labelling means heavy labour involvement, thus, produces highly expensive yet low-quality supervision. To make matters worse, since HPIM is based on the attribute describer trained on $\overline{X}$, which should robustly adapt to the Re-ID dataset with a possible domain shift. Finally, attribute imbalance commonly occurs in the large-scale person collection. As shown in Fig. \ref{dataDistribution}, the popular person attribute dataset PETA has shown an extreme case that the data distribution of each attribute is highly unbalanced. This phenomenon further brings a new challenge to obtain trustful multi-attribute description codes. The difficulties drive us to re-process the attribute set by the grouping technique we are about to mention. Correspondingly we further invent a multi-label weight-shifting training procedure through semi-supervised learning, showing the ability to resist the corrupted-labelled training examples.

\textbf{Person attribute grouping.}
Human attributes are correlated with each other. For example, people with \emph{long hair}, \emph{red skirt} is prone to a \emph{woman}, and a man with \emph{down jacket}, \emph{cotton shoes} barely wears a \emph{T-shirt}. This trait inspires a new person attribute reformation we name \emph{attribute grouping}. Concretely, suppose there are $M$ attributes for each example, e.g.,$\{a_1,a_2,\cdots,a_M\}$, we then define $G$ attribute groups as $\mathcal{G}_1, \mathcal{G}_2, ... , \mathcal{G}_G$, each of which represents a subset of attributes semantically exclusive with each other and meanwhile, $\forall i,j\in[G]$, $\mathcal{G}_i\cap \mathcal{G}_j=\emptyset$. For instance, the group ``shoes" contains attributes \emph{sandals}, \emph{sneaker}, \emph{leather shoes, et al}. They are spatially unique, thus, can be treated as a set of exclusive categories. Besides, to ensure the compactness among the attribute classes of each group, we add the group-specific negative attribute $\hat{a}^{(i)}$ to the group $\mathcal{G}_i$, which annotates the examples not belonging to any attributes in the group. The reforming implicitly refers to $G$ multi-class recognition problems, which is able to reduce the task complexity (a multi-label classification from $M$ to $G$ that $M$ is mostly bigger than $G$). 

%Moreover, as the comparison between the statistic before and after the grouping in Fig., the variance of image number across attributes has also been significantly reduced. It reveals the positive effect of mitigating the category imbalance.

Formally, given $\forall i\in [L]$, $\mathcal{G}_i$ denotes a group with $M_i+1$ exclusive attributes, namely, $\mathcal{G}_i=\{a^{(i)}_1, \cdots, a^{(i)}_{M_i}, \hat{a}^{(i)}\}$ and $\forall i\in[G]$, $\{a^{(i)}_1, \cdots, a^{(i)}_{M_i}\}\subseteq \{a_1,a_2,\cdots,a_M\}$ with $\overset{L}{\underset{i}{\cup}}\{a^{(i)}_1, \cdots, a^{(i)}_{M_i}\}=\{a_1,a_2,\cdots,a_M\}$. Then image $\overline{x}\in\overline{X}$ has been annotated with $G$ attribute group labels $(y^{(1)}, \cdots, y^{(G)})$, where $y^{(i)}$ denotes an one-hot vector with $M_i+1$ dimensions. Our describer is born as a $G$-label classifier learning through shifting weight.

\textbf{Attribute describer learning.} After redefining the attribution recognition problem, we employ a deep neural network $h_{\theta_{d}}(\cdot)$ as a backbone of our describer proposed for the $G$-label classification under the noisy training environment with $\overline{X}$. Specifically, suppose each image $\overline{x}_i\in\overline{X}$ has been annotated as $(\overline{x}_i, y^{(1)}_i, \cdots, y^{(G)}_i)$, where $\forall j\in[G]$, $y^{(j)}_i$ denotes the grouped attributes of $\mathcal{G}_j$ about the sample $\overline{x}_i$. We provide a softmax classification $h^{(j)}_{\theta_d}(\overline{x}_i)$ based on the group $j$ to organize the $G$-label visual recognition, 
\begin{equation}\label{eq3.1}
\begin{aligned}
  \underset{\theta_d}{\min}\frac{1}{|\overline{X}|}\sum_{i=1}^{|\overline{X}|}\sum_{j=1}^{G} \mathcal{L}(h^{(j)}_{\theta_d}(\overline{x}_i),y^{(j)}_i)
\end{aligned}
\end{equation}
where $\mathcal{L}(h^{(j)}_{\theta_d}(\overline{x}_i),y^{(l)}_i)$ denotes the cross entropy loss specific for the $j^{th}$ group. But in order to overcome the noisy images, we prefer \emph{shifting weight}, a self-paced learning based training technology to resist the corrupted labels and mild domain shifts:  
\begin{equation}\begin{aligned}\label{eq3.2}
\underset{\mathbf{v}, \theta_d}{\min} \sum_{i=1}^{\overline{X}} \sum_{j=1}^{G} v^{(j)}_{i} \mathcal{L} (h^{(j)}_{\theta_d}(\overline{x}_i), y_i^{(j)}) -\lambda_j v^{(j)}_{i}
\end{aligned}
\end{equation}
where $v^{(j)}_i$ denotes the binary latent weight variable of training example $\overline{x}_i$ referred to the group $j$ and $\mathbf{v}=\{v^{(j)}_i\}^{|\overline{X}|,G}_{i=1,j=1}$. Then Eq. \ref{eq3.2} turns to an EM-like learning objective that before optimizing the describer parameter $\theta_d$, we need to infer $\mathbf{v}$ based on the previous describer. Obviously given a fixed $\theta_d$, Eq. \ref{eq3.2} w.r.t. $\mathbf{v}$ is convex with a following close-form solution,  
\begin{equation}\label{spl}
v^{(j)}_i =
\begin{cases}
1, & \text{$ \mathcal{L} (h_{\theta_d}(\overline{x}_i), y_i^{(j)}) < \lambda_j, $} \\
0, & \text{otherwise}.
\end{cases}
\end{equation}
where $\lambda_j$ is an auto-decisive threshold to accept or reject each sample to join the next-step optimization of the $j^{th}$ attribute group classification. Following the vanilla self-paced manner, $\lambda_j$ starts with a value merely covering half of $\overline{X}$, then progressively grows larger to receive more examples with bigger losses (In our experiments, we adopt the increasing rule as \cite{Kumar2010Self}). The strategy tends to select the trustful training samples first, then move to the examples more possible to be outliers. It empirically performs excellent to protect the model from the corrupted label contamination.  

\begin{figure}[t]
	\centering
	\includegraphics[width=4in]{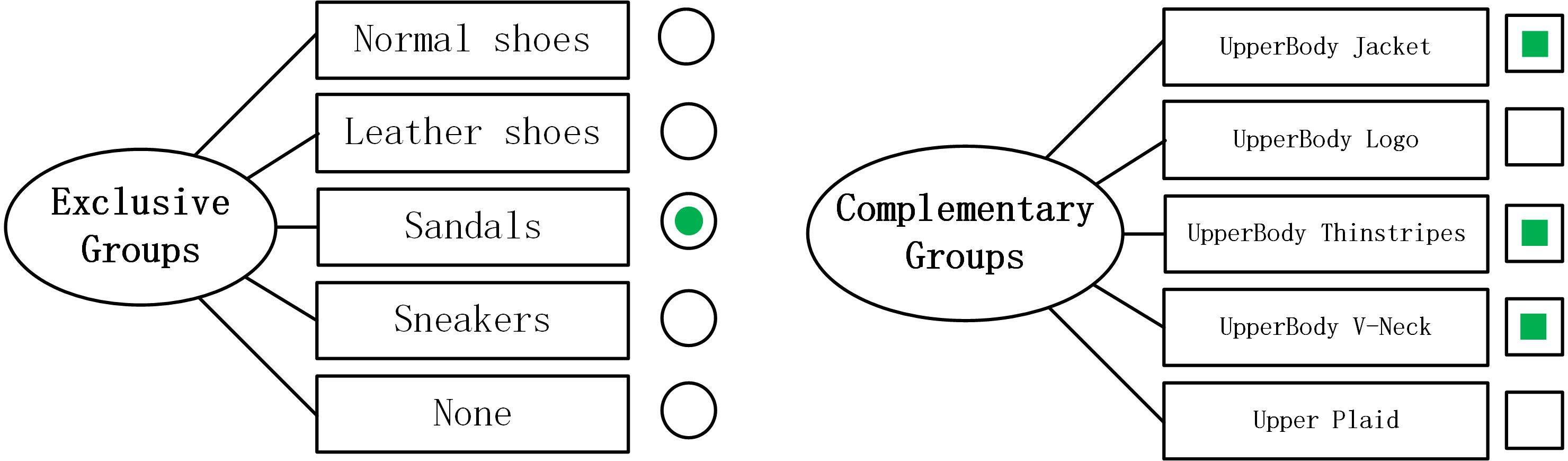}
	\caption{ The illustration of our semantic attribute grouping. The left sub-figure is an exclusive groups (\emph{shoes} for an example); the right sub-figure is a complementary groups (\emph{upper body dress} for an example). }
	\label{attribute_Groups}
\end{figure}	

%% ================================================================
%% ================	Group-driven Self-Paced Learning	===========
%% ================================================================

%Formally, we utilize $x_i$ to denote the \emph{$i^{th}$} labeled data, and use $(x_i,y_i^{G_g})$ to represent corresponding group $g$'s attribute labels. We first use the labeled human images to train a deep neural network and obtain the pre-trained CNN. After feeding the pre-trained deep model with plenty of sampled unlabeled data, we can obtain the predicted score of these samples. Hence, the group-based cross entropy loss function (\emph{attribute score}) $\mathcal{L}_g$ can be represented as:

%The overall algorithm of our group-wise self-paced pedestrian attributes recognition framework can be found in Algorithm \ref{SPLAttributesAlgorithm}. 

\subsection{\textbf{Stage Two: Hard Person Identity Mining}} \label{PersonREIDHIM}
The outcome of the previous stage is a well-trained attribute describer $h_{\theta^\ast_d}(\cdot)$. In our re-ID setup with identity set $\{X_j\}^N_{j=1}$, given an image $x^{(j)}_i\in X_j$ ($\forall j\in[N]$), the output $h_{\theta^\ast_d}(x^{(j)}_i)$\footnote{$h_{\theta^\ast_d}(x^{(j)}_i)$ is concatenated by $\{h^{(1)}_{\theta^\ast_d}(x^{(j)}_i),\cdots,h^{(G)}_{\theta^\ast_d}(x^{(j)}_i)\}$.} denotes the description code of $x^{(j)}_i$. Since the describer is only applied for encoding images, we rewrite the description codes as $c^{(j)}_i=h_{\theta^\ast_d}(x^{(j)}_i)$ for simplicity.

It is observed that given an identity $j$, the codes $C_j=\{c^{(j)}_i\}$ only refer to the images with the same identity agreeing with the attributes. We assume that, each code set $C_i$ underlies a latent distribution $P_i$, deriving the same visual appearance of the $i^{th}$ person. In this consideration, more similar distributions in $\{P_i\}^N_{i=1}$ indicates the persons sharing more semantic attributes, thus, harder to be told apart. 

The key idea is to assemble training batches by discovering hard identities. In specific, we measure the statistical discrepancy among $\{P_i\}^N_{i=1}$. The small value means the related person identities close in attributes, thus, organizes the hard identities during training. Though the target is to calculate the differences among $\{P_i\}$, whereas we know nothing except for $\{C_j\}^N_{j=1}$ from the distributions. To attain the statistic information, we introduce Central Moment Discrepancy (CMD) as the nonparametric measurement to estimate the distances between these latent distributions.       

\textbf{Central Moment Discrepancy (CMD).} Before further explaining our method, we provide an overview of CMD underlying our re-ID setup. The CMD terms mostly mentioned in the paper points to the empirical estimation \cite{egozcue2012smallest} of the CMD metric \cite{zellinger2017central, zellinger2019robust}. 

[\textbf{(CMD metric)}]
Let $C_j = (c^{(j)}_1,\cdots, c^{(j)}_n )$ and $C_k = (c^{(k)}_1,\cdots, c^{(k)}_n)$ be bounded random vectors independent and identically distributed from two probability distributions $P_k$ and $P_j$ on the compact interval $[0,1]^{M+G}$. The central moment discrepancy metric (CMD metric) is defined by: 
\begin{equation}\label{cmdm}
	CMD(P_i,P_j)=||\mathbb{E}(C_k)-\mathbb{E}(C_j)||_2+\sum_{l=2}^{\infty}||m_l(C_k)-m_l(C_j)||_2
\end{equation}
where $\mathbb{E}(X_j)$ and $\mathbb{E}(X_k)$ denote the expectation of $X_j$ and $X_k$ respectively, and 
\begin{equation}
m_l(X)= \Big(\mathbb{E}\big(\overset{M+G}{\prod}(c_i-\mathbb{E}(c_i))^{r_i}\big)\Big)_{\sum_{i=1}^{M+G}r_i=l, r_1,\cdots,r_{M+G}>0}
\end{equation} 
indicates the central moment vector of order $l$. The first order central moments are means, the second order central moments are variances, and the third and fourth order's moment respectively denote the skewness and the kurtosis. The theoretical form of CMD metric (Eq. \ref{cmdm}) is not calculable due to the infinity term, and in practice, we turn to its empirical estimation:
[\textbf{(CMD)}]
	Let $C_j = (c^{(j)}_1,\cdots, c^{(j)}_n)$ and $C_k = (c^{(k)}_1,\cdots, c^{(k)}_n )$ be bounded random
	vectors independent and identically distributed from two probability distributions $P_j$ and $P_k$ on the
	compact interval $[0,1]^{M+G}$. $CMD_{L}$ denotes a $L$-order empirical estimate of the CMD metric, by
	\begin{equation}\label{cmd}
	CMD_{L}(C_j,C_k)=||\mathbb{E}(C_j)-\mathbb{E}(C_k)||_2+\sum_{l=2}^{L}||M_l(C_j)-M_l(C_k)||_2
	\end{equation}
where $M_k(C)= \mathbb{E}\big((c_i-\mathbb{E}(C))^k\big)$. Hence given identity pair $j,k$, through calculating $CMD_{L}(C_j,C_k)$ we are able to obtain an approximated difference between $P_j$ and $P_k$.

\textbf{CMD-based stochastic identity sampling.} Based on the previous discussion, we are capable to quantify all the CMD-based relationships among identities regardless of the evolving embedding space. But like sample mining approaches mostly appealing, hard identity mining should be able to balance the exploration and exploitation during discovering hard identity. We formulate a $K$-nearest-neighbor driven stochastic sampling policy for implementing the CMD-based algorithm to achieve this goal.  

Specifically, suppose $a$ is the anchor identity we randomly consider and $Knn(a)$ denotes a set of identities with the top-$K$ smallest CMD values with the anchor. We kernelize the CMD values, namely, 
\begin{equation}\label{kernel}
h(a,j) = exp\Big(-\frac{\big(CMD_L(C_a,C_j)\big)^2}{\sigma^2}\Big), \ \forall j\in [N]/{a}
\end{equation}where $\sigma>0$ denotes the kernel's bandwidth. The hard identity with more similar appearances owns smaller CMD value, thus, larger $h(a,j)$. Then provided $Pro(j;a)$ as the probability of choosing identity $j$, we define a stochastic policy sampling as follow: 
\begin{small}
\begin{equation}\label{policy}
Pro(j;a) =
\begin{cases}
\frac{h(a,j)}{\underset{i\in [N]/\{a\}}{\sum} h(a,i)} &  j\in Knn(a)\\
\frac{1}{|[N]/Knn(a)\cup\{a\}|}*\Big(1-\frac{\underset{i\in Knn(a)}{\sum} h(a,i)}{\underset{i\in [N]/\{a\}}{\sum} h(a,i)}\Big) & otherwise\\
\end{cases}
\end{equation}\end{small}
As we see from Eq. \ref{policy}, the candidate identity $j$ belonging to the anchor's $k$ nearest neighbors, perceives the selection probability $\frac{h(a,j)}{\underset{i\in [N]/\{a\}}{\sum} h(a,i)}$ via exploiting CMD with the anchor. The closer neighbor ID has larger kernalized CMD value and obtains a higher selection probability. If the candidate identity does not belong to the anchor's neighbors, they share the same exploring probability to be selected, which also relates to the top-$k$ CMD values. Particularly compared with the other identities, if the neighbor identities appear very close to the anchor, the numerator $\underset{i\in Knn(a)}{\sum} h(a,i)$ will dominate the denominator $\underset{i\in [N]/\{a\}}{\sum} h(\rho(a,i))$ and we have a observation below: 

[\textbf{HPIM exploitation adjustment}]\label{prop1}
Given an anchor identity $a$ and its $K$ nearest neighbor identity set $Knn(a)$, if $\forall j\in Knn(a)$ and $\forall i\in[N]/Knn(a)\cup\{a\}$ there exist $CMD_L(X_i,X_j)\rightarrow +\infty$, we have
	\begin{equation}
	Pro(j;a) =
	\begin{cases}
	\frac{h(a,j)}{\underset{i\in Knn(a)}{\sum} h(a,i)} &  j\in Knn(a)\\
	0 & otherwise\\
	\end{cases}
	\end{equation}

\begin{proof}
	Due to the nearest neighboring, we assume the upperbound of CMD between the anchor and neighbors is $B$: 
	\begin{equation}
	\forall j\in Knn(a), \ \ 0<CMD_L(C_a,C_j)\leq B. 
	\end{equation}
	Since $CMD_L(\cdot,\cdot)$ is a metric, by triangle inequality, we have a simple speculation,
	\begin{displaymath}\begin{aligned}
	CMD_L(C_i,C_j)-CMD_L(C_a,C_j)\leq CMD_L(C_i,C_a)\\
	CMD_L(C_i,C_j)\leq CMD_L(C_i,C_a), 
	\end{aligned}
	\end{displaymath}
	then $CMD_L(C_i,C_a)\rightarrow +\infty$ and $h(a,i)=0$. Therefore,
	\begin{displaymath}
	\begin{aligned}
	\frac{h(a,j)}{\underset{i\in [N]/\{a\}}{\sum} h(a,i)}&=\frac{h(a,j)}{\underset{i\in [N]/Knn(a)\cup\{a\}}{\sum} h(a,i)+\underset{i\in Knn(a)}{\sum} h(a,i)} \\&=\frac{h(a,j)}{\underset{i\in Knn(a)}{\sum} h(a,i)}, 
	\end{aligned}
	\end{displaymath}
	and
	\begin{displaymath}\begin{aligned}
	1-\frac{\underset{i\in Knn(a)}{\sum} h(a,i)}{\underset{i\in [N]/\{a\}}{\sum} h(\rho(a,i))}&=1-\underset{i\in Knn(a)}{\sum}\frac{ h(a,i)}{\underset{i\in [N]/\{a\}}{\sum} h(\rho(a,i))}\\&= 1-\underset{i\in Knn(a)}{\sum}\frac{ h(a,i)}{\underset{i\in Knn(a)}{\sum} h(\rho(a,i))}\\&=0. 
	\end{aligned}
	\end{displaymath}Conclude the proof. 
\end{proof}
It means that if the CMD-based $k$ nearest neighbours are relatively trustful, the exploring scheme triggered by probability $1-\frac{\underset{i\in Knn(a)}{\sum} h(a,i)}{\underset{i\in [N]/\{a\}}{\sum} h(\rho(a,i))}$ will be possibly abandoned. On the other hand, suppose the anchor identity contains so many neighbours close to it. It maintains,
[\textbf{HPIM exploration adjustment}]\label{prop2}
	Given an anchor identity and if $\forall j\in Knn(a)$, $CMD_L(C_a,C_j)\rightarrow 0$, and $\forall i\in [N]/Knn(a)\cup\{a\}$ $\underset{j\in Knn(a)}{\min} CMD_L(C_j,C_i)\leq\mu$. Then given $\mu\rightarrow 0$, 
	\begin{equation}\label{15}
	Pro(j;a)=\frac{1}{N-1}, \ \forall j\in[N]/\{a\}
	\end{equation}

\begin{proof}$\forall j\in Knn(a)$ it is obvious that $CMD_L(C_a,C_i)\leq CMD_L(C_a,C_j)+CMD(C_j,C_i)$ ($j\in Knn(a)$). We choose $j^\ast=\arg\underset{j\in Knn(a)}{\min} CMD_L(C_j,C_i)$, so $CMD_L(C_a,C_i)\leq CMD_L(C_a,C_j)+\mu$, and $CMD_L(C_a,C_j)\rightarrow 0$, $\mu\rightarrow 0$ leads to $CMD_L(C_a,C_i)\rightarrow 0$. Hence $\forall j\in [N]/{a}$,
\begin{equation}
h(a,i) = exp\Big(-\frac{\big(CMD_L(C_a,C_j)\big)^2}{\sigma^2}\Big)=1 
\end{equation} that results in Eq. \ref{15}.
\end{proof} 
Proposition. \ref{prop2} demonstrates a fact: if the neighbour IDs are sufficiently close to the anchor IDs and other IDs also perform close to the neighbours, HPIM develop to simulate a random selection strategy for mining hard identity.

In the training phase, we argue that the distance between each sample and its averaged sample can be viewed as a criterion to training the neural network. We think the person images with same ID should be clustered as close as possible, that is to say, each sample should be close to their average samples. According to this assumption, we add a regularization term into the triplet loss function and detailed information can be described as follows. For each ID $i$ in one mini-batch, we can obtain it's average attribute $\overline{r}^i$ and average feature $f_\theta(\overline{x}^i)$ as follows:
\begin{equation}
\begin{aligned}
\overline{r}^i &= \frac{1}{m}\sum_{j}^{m}r_j^i , ~~~ f_{\theta}(\overline{x}^i) = \frac{1}{m}\sum_{j}^{m}f_{\theta}(x_j^i)
\end{aligned}
\end{equation}

For each image $a$ in corresponding ID $i$, we have loss function described as follows:
%\begin{equation}
%\begin{aligned}
%\mathcal{L}_s(x_a^i) = m \ + \ \underset{p=1,\cdots,K}{\max} \ell(f_{\theta}(x^i_a),f_{\theta}(x^i_p)) \ - \underset{
%\substack{j=1,\cdots,P \\ n = 1,\cdots,K \\ j\neq i}}{\min} \ell(f_{\theta}(x^i_a),f_{\theta}(x^j_n))
%\end{aligned}
%\end{equation}
\begin{equation}\label{eq1}
\begin{small}
\begin{aligned}
\mathcal{L}_s(x_a^i)  = \sum_{i=1}^{P}\sum_{a=1}^{K}[m \ + \ \underset{p=1,\cdots,K}{\max} \ell(f_{\theta}(x^i_a),f_{\theta}(x^i_p)) \ -   \underset{
	\substack{j=1,\cdots,P \\ n = 1,\cdots,K \\ j\neq i}}{\min} \ell(f_{\theta}(x^i_a),f_{\theta}(x^j_n))]
\end{aligned}
\end{small}
\end{equation}
Here, where $f_{\theta}(x)$ denotes the feature of person image $x$ extracted from neural networks with parameter $\theta$. $P$ is the number of person IDs we use in each mini-batch, $K$ is the number of images for each ID in each mini-batch. $m$ denotes the margin used in the loss function and it is a scalar value. $\ell$ is a loss function and we use soft margin loss as below:
\begin{equation}
\begin{aligned}
\ell(x,y) &= \log(1+e^{d(x,y)})
%\ell(x,y,\alpha) &= \log(1+e^{\alpha d(x,y)})
\end{aligned}
\end{equation}
where $d$ is the Euclidean distance. Therefore, the improved triplet loss function can be described as:
\begin{equation}\label{2}\begin{aligned}
\mathcal{L}(x_a^i) = \mathcal{L}_s(x_a^i) + \alpha\ell(f_{\theta}(x^i_a),f_{\theta}(\overline{x}^i))
\end{aligned}
\end{equation}
where $\alpha$ is a trade-parameter, we set it as $0.55$ in all our experiments.

Based on previous discussions, we summarize the whole learning framework in Algorithm. \ref{Algorithm}. 

\renewcommand{\algorithmicrequire}{\textbf{Input:}}
\renewcommand{\algorithmicensure}{\textbf{Output:}}
\begin{algorithm}[tb]
	\caption{MEL with Improved Hard Example Ming by Hard Person Identity Mining}
	\label{Algorithm}
	\begin{algorithmic}[1]
		\REQUIRE 
			Re-ID mapper $f_{\theta}$, attribute describer $h_{\theta_d}$, person identity set $\{X_1,\cdots,X_N\}$, person attribute set $\overline{X}$.\\
		\ENSURE    
		
		\STATE \textbf{Stage 1:}
		\STATE Divide the attributes into $G$ groups by attribute grouping, then $\forall \overline{x}\in\overline{X}$, its supervision is $(y^1,\cdots,y^{G})$.
		\WHILE{\emph{not converged}}
		\STATE Update $\theta_d$ by solving Eq. \ref{eq3.2} through stochastic training
	%	\STATE  Input sampled data $\mathcal{U}_{sub}$ from $\mathcal{D}_U$ to pre-trained CNN, obtain the predicted scores of each sample;
	%	\STATE Assign labels according to predicted score and \emph{group prior};
		\FOR{each group $j\in[G]$}
		%\STATE Sort the samples in ascending order of their cross-entropy loss values computed via Eq. (\ref{lossFunction});
		
		\FOR{each sample i=1 to $n$ }
		\STATE Update $v^{j}_i$ by Eq. \ref{spl}.
		\ENDFOR
		\ENDFOR
		
		\STATE Update $ \{\lambda_1,\cdots,\lambda_G\}$ by $\mu$;
		\ENDWHILE
		\STATE Obtain $\theta^\ast_d$.
	    \STATE \textbf{Stage 2:}
	    \STATE Obtain attribute description codes for $\{X_1,\cdots,X_N\}$:
	    \FOR{$j=1$ to $N$}
	    \FOR{$i=1$ to $K$}
	    \STATE $c^{(j)}_i=[h^{(1)}_{\theta^\ast_d}(x^{(j)}_i),\cdots,h^{(G)}_{\theta^\ast_d}(x^{(j)}_i)]$,
	    \ENDFOR \ \ $C_j=\{c^{(j)}_i\}$;
	    \ENDFOR
	    \FOR{$a$,$j=1$ to $N$}
	    \STATE Calculate $CMD_L(C_a,C_j)$ by Eq. \ref{cmdm}
	    \STATE Obtain $h(a,j)$ by Eq. \ref{kernel}
	    \ENDFOR 
	    \WHILE{\emph{not reach the iteration $T$}}
	    \STATE Sample $P$ identities by Eq. \ref{policy} and use these identities to form a training batch;
	    \STATE Update $\theta$ in Eq. \ref{eq2.1} / \ref{eq2.2} with the training batch;
	    \ENDWHILE
	\end{algorithmic}
\end{algorithm}

%========================================================================================================
%=========================================  			Experiments  			 =========================================
%========================================================================================================
\section{Experiments}\label{experiments}
To validate the effectiveness of our proposed method, including group-wise self-paced attribute recognition and hard person identity mining for person re-identification, we validate these two tasks on public benchmarks, respectively. For pedestrian attribute recognition, we implement the experiments on public attribute dataset (PETA dataset \cite{deng2014pedestrian}), and also introduce unlabelled human images to further validate the effectiveness of our proposed group-wise self-paced attribute recognition approach. For the person re-identification, the experiments are implemented on two popular benchmarks, \emph{i.e.} Market-1501 \cite{zheng2015scalable} and CUHK03 \cite{li2014deepreid}.

Specifically, we first introduce the evaluation criterion for both tasks in section \ref{evaluationCriterion}; then, the introduced dataset and related parameter settings are given in section \ref{datasetDescriptionPS}. Section \ref{PETAresults} mainly focus on the experimental results on PETA dataset, and ablation studies on group-wise self-paced learning. Section \ref{personreIDMarket1501CUHK03} is utilized to report the re-identification performance of our methods and other state-of-the-art algorithms on two public person re-ID benchmarks.

%% --------------------------- Experimental Setup and Parameter Setting ------------------------------
\subsection{Evaluation Metric}\label{evaluationCriterion}
For person attribute recognition, we compare our results with PETA benchmark baselines and other state-of-the-art approaches via the following evaluation criterion:
\begin{equation}
\label{accuracy}
Accuracy = 0.5 * (\frac{TP}{TP+FN} + \frac{TN}{TN+FP})
\end{equation}
where $TP, TN, FP, FN$ denotes \emph{true positive}, \emph{true negative}, \emph{false positive} and \emph{false negative}, respectively.

For person re-identification, mean Average Precision (mAP) is a popular criterion to measure the re-ID perfromance. For each query, its average precision (AP) is computed from its precision-recall curve. Then mAP is the mean value of average precisions across all queries. Another evaluation metrics is the Cumulative Matching Characteristic (CMC), and the CMC reflects retrieval precision, while mAP reflects the recall.

\subsection{Dataset Description and Implement Details}\label{datasetDescriptionPS}
%% -------------------------------- PETA dataset

\begin{figure}[t]
\center
\includegraphics[width= 3.5in]{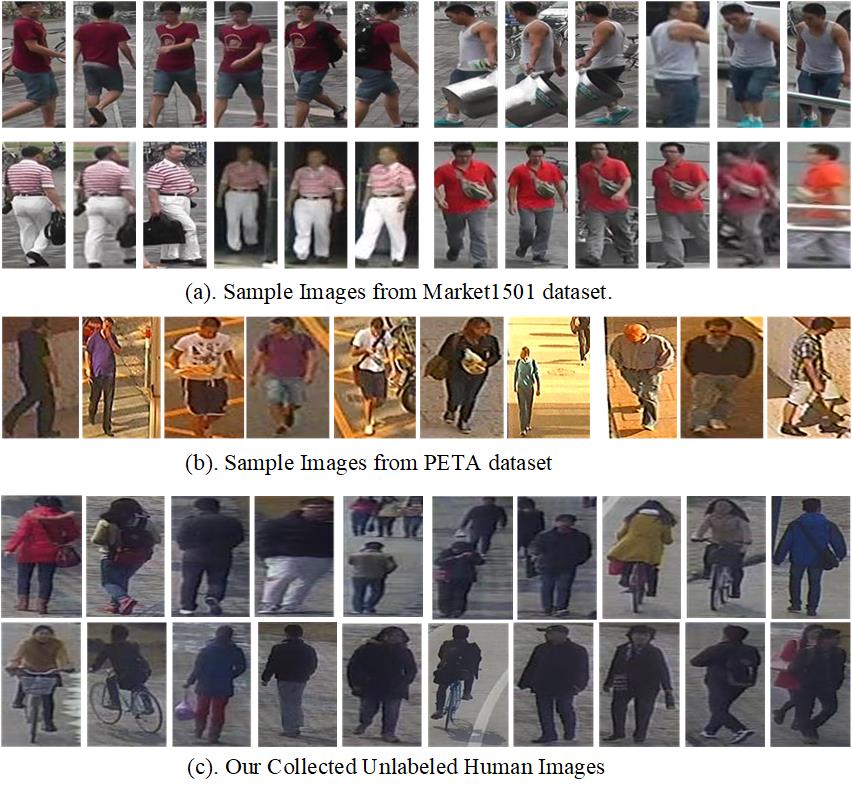}
\caption{Sample images of pedestrian attribute dataset (PETA dataset) and person re-identification benchmark (Market1501 and CUHK03) utilized in this paper.}
\label{sampleAttributesimages}
\end{figure}

\begin{table*}[htp!]
\center
\scriptsize
\caption{Comparison of our results with other state-of-the-art algorithms on the PETA dataset.}\label{result-PETA}
\begin{tabular}{l|ccccccccccccr}
\hline
\hline		
\textbf{Algorithms}   	   &ikSVM \cite{deng2014pedestrian} &MRFr2-fw \cite{deng2014pedestrian} &DeepSAR \cite{deng2014pedestrian}  &DeepMAR \cite{deng2014pedestrian} &LP \cite{Xiaojin2005Semi}&DLP \cite{Wang2013Dynamic} 	&GRL \cite{ijcai2018GRL} \\
\hline															
\textbf{Average} 			&69.5 	&75.6	&81.3	&82.6			  &74.9		&61.7	&84.34	\\
\hline	
\textbf{Algorithms}   	 &HydraPlus-Net \cite{liu2017hydraplus} 	   &ACN \cite{sudowe2015personworkshop} &CTX CNN-RNN \cite{li2017sequential} &SR CNN-RNN \cite{liu2017semantic} &JRL \cite{Wang2017Attribute}   &AlexNet (baseline)  &Ours  \\
\hline															
\textbf{Average} 		&84.92		&84.06 &79.68 &82.54 &86.03 &83.0  &84.9			\\
\hline	
\end{tabular}
\end{table*}

\begin{table}[htp!]
\center
\caption{Compare results of different annotation usage on PETA dataset. }\label{analysisPETA}
\begin{tabular}{l|ccccccccccccr}
\hline
\hline
Initial  					&30\%  &30\%  &50\% &50\%  &100\% \\
\hline
Self-annotated 			&0\% & 70\% & 0\% & 50\% & 0\% \\
\hline																	
Average 				&76.4		&81.1		&79.1		&81.8		&83.0	  		\\
\hline	
\end{tabular}
\end{table}

\textbf{PEdesTrian Attribute (PETA) dataset} \cite{deng2014pedestrian} is a very challenging dataset, which contains $19,000$ pedestrian images with different \emph{camera angles, view point, illumination, resolution} and includes \emph{indoor} and \emph{ outdoor} scenarios. For fair comparison, we strictly follow the same protocol in \cite{deng2014pedestrian}, \emph{i.e.}, we randomly select $9,500$ images as training data, $1,900$ for verification, and $7,600$ for testing. We select out the same $35$ pedestrian attributes as this benchmark does and compare the recognition accuracy with other baseline approaches based on these attributes. These attributes are very representative. They consist of $15$ most important attributes in video surveillance and other $20$ attributes covering all body parts of pedestrian. The specific attributes could be found in Table \ref{result-PETA}.

\textbf{Our Collected Unlabelled Images.} To fully evaluate the effectiveness of our proposed  group-wise self-paced learning, we not only verify our method within public pedestrian attributes dataset, but also on the additional collected unlabelled data. About $30,000$ human images from monitoring scenarios videos are collected as unlabelled data used in this paper. Sample images of person attributes from PETA and our collected images are shown in Fig. \ref{sampleAttributesimages}. These data can be found at: \url{https://sites.google.com/view/attrpersonreid/}. 

\textbf{The Market1501 Dataset} \cite{zheng2015scalable} is one of the largest person re-ID datasets, contains 32,668 gallery images and 3,368 query images captured by 6 cameras. It also includes 500k irrelevant images forming a distractor set, which may exert a considerable influence on the recognition accuracy. Market-1501 is split into 751 identities for training and 750 identities for testing.

\textbf{The CUHK03 Dataset} \cite{li2014deepreid} contains 1467 identities and 28192 BBox which is obtained from detection algorithm DPM or manual annotated. Based on the benchmark, 26264 BBox of 1367 identities are used for training and 1928 BBox of 100 identities are utilized for testing.

\textbf{Implement Details.}
For the experiments of human attributes recognition, AlexNet~\cite{Krivsky2012ImageNet} is adopted as our basic deep architecture and build upon a sigmoid layer behind the \emph{fc8} layer. The parameters of our initial learning rate is 0.001, momentum equal to 0.9 and weight decay is 0.0005. For the person re-ID, we utilize the Residual-50 network \cite{He2015Deep} as our basic network architecture and triplet-loss as our basic loss function. Our implementation for person re-identification is based on the following project {\url{https://github.com/Cysu/open-reid}}.
	
% In the training phase, we argue that the distance between each sample embedding and the averaged embedding with its identity can be viewed as a criterion to training the neural network. Since the person images with the same ID should be clustered together, that is to say, each sample should be close to their average embedding. We add a regularization to achieve this goal, and the balance factor is set as $0.55$ in all the experiments.

%% --------------------------------------------------- PETA dataset
\subsection{Experiments on PETA dataset}\label{PETAresults}
	We will first analyse the recognition results obtained by each methods on PETA dataset  \cite{deng2014pedestrian}. As shown in Table~\ref{result-PETA} , we can find that the accuracy of our approach, which are produced by joint training of CNN models, are obviously better than the baselines of Deng \emph{et al.} provided. More specifically, our accuracy surpass their baseline more than $10\%$, and most attributes are better, either. We can obtain from Table \ref{result-PETA} that the accuracy of ikSVM only have $69.5\%$, due to the challenging pedestrian images. Some graph models could build the relationships between many elements such as CRF, MRF \emph{et al.} and has been widely used in many computer vision tasks. Deng \emph{et al.} also introduce MRF into attribute recognition to build the relationships between human images,  and increase the accuracy from $69.5\%$ to $71.1\%$. They also combine the feature of foreground human and background or whole images to better represent the attribute, however, their accuracy, \emph{i.e.} $75.6\%$, is still lower than deep learning approach as shown in MRFr2-fw. Li \emph{et al.} improved the performance to $81.3\%$ via training one CNN model for one attribute. However, as we can see from the Table \ref{result-PETA}, joint training the attribute can learning the potential relations between these attributes and obtain better results \emph{i.e.} DeepMAR obtained $82.6\%$ via joint training with a new designed loss function. We achieve a comparable accuracy \emph{i.e.} $83.0\%$ when all the train data are used for finetuning AlexNet~\cite{Krizhevsky2012ImageNet} based on deep models pre-trained based on ImageNet \footnote{\url{http://caffe.berkeleyvision.org/model\_zoo.html\#bvlc-model-license}}.
	
In addition, we also compare our algorithm with some graph-based semi-supervised learning methods, which are simple but rather effective, such as LP \cite{Xiaojin2005Semi}, DLP \cite{Wang2013Dynamic}. As shown in Table \ref{result-PETA}, those two methods are all semi-supervised learning approach which utilize 50\% train data to pre-train a CNN model and extract the deep feature to feed those algorithms to propagate the labels from labelled data to unlabelled data. The propagated unlabelled data are mixed with original train data to finetune the CNN model further. However, those method don't take the relationships between labels and the difficulty of samples into consideration. And as a common problem of semi-supervised learning methods, the easily introduced noise in each iteration may also limit the improvement of recognition accuracy. 
	
We take the correlations between pedestrian attributes into consideration and learn these unlabeled data in a self-paced manner based on introduced semantic attribute groups, which further improve the recognition accuracy. As we can see in Table \ref{analysisPETA}, with $50\%$ train data, we can utilize the rest unlabelled data improve from $79.1\%$ to $81.8\%$. As demonstrated in Table \ref{result-PETA}, with $100\%$ train data used, we can achieve $83.0\%$ on this dataset. With the help of our collected unlabelled data, we can improve this baseline from $83.0\%$ to $84.9\%$, this is a significant improvement compared with other methods, such as: ACN \cite{sudowe2015personworkshop}, CTX CNN-RNN \cite{li2017sequential}, SR CNN-RNN \cite{liu2017semantic}. We also achieve comparable or even better results than recent person attribute recognition algorithm, such as GRL \cite{ijcai2018GRL}, HydraPlus-Net  \cite{liu2017hydraplus}.

In addition, we also implement experiments on PETA dataset to test the minimal usage of train data, and try to achieve comparable results with fully-supervised approaches.  Thus, we divide the original train data into labelled subset and unlabelled subset according to percent of train data, $30\%$, $50\%$, respectively. As we can see, our semi-supervised learning approach improved the accuracy from 76.4\% to 81.1\% on 30\% used, and from 79.1\% to 81.8\% on 50\% selected labelled data, respectively. It is also worthy to denote that this performance achieve comparable results with existing fully-supervised deep learning methods and even beyond some benchmark baselines. These experimental results all demonstrate the effectiveness of our proposed group-wise self-paced learning in dealing with unlabelled data.

According to the experimental results on public pedestrian attribute benchmark and external collected unlabelled data, we can find that our framework is indeed effective and achieve good recognition results compared with other approaches.

\subsection{Experiments on Market1501 and CUHK03 dataset}\label{personreIDMarket1501CUHK03}
Two popular person re-ID benchmarks (\emph{i.e.} Market1501 and CUHK03) are selected to evaluate the effectiveness of our proposed hard person identity mining algorithm. We will introduce the comparisons between our method and other person re-ID approaches and  ablation studies in following subsections, respectively.

%	\textbf{Comparison with the state-of-the-art methods.}
The comparison with the state-of-the-art algorithms on Market1501 and CUHK03 dataset is shown in Table \ref{PersonREIDMarket1501} and Table \ref{PersonREIDcuhk03benchmark}, respectively. On the Market1501 dataset, we obtain rank-1 = 79.6\%, mAP = 62.2\% using the ResNet-50 model and 751 training IDs. We achieve the best rank-1, rank-5, rank-10 and mAP accuracy among all mentioned methods. On the CUHK03 benchmark, our results are rank-1 = 64.7\%, rank-5 = 88.9\% and rank-10 = 93.7\%. It is obvious that our method achieve the best rank-1 accuracy and second best rank-5 and rank-10 performance among the competing methods. Hence, we can find that our proposed hard person identity mining can help person re-identification achieving favorablely with the state-of-the-art methods.

\begin{table}[htp!]
\center
\caption{Person Re-ID Results on Market-1501 Benchmark.}\label{PersonREIDMarket1501}
\begin{tabular}{l|llll}
\hline
\hline
Methods  		&top-1  &top-5  &top-10 	& Mean AP\\
\hline
DADM  \cite{Su2016Deep} 	   			&39.4		&-		&-		&19.6	\\					
GAN  \cite{ICCV2017Unlabeled}			 		&79.33		&-		&-		&55.95 		\\												
MBC   \cite{Ustinova2017Multi}			 		&45.56		&67		&76		&26.11 		\\															
SML  \cite{Jose2016Scalable}			 		&45.16		&68.12		&76		&- 		\\																
DLDA  \cite{Wu2017Deep}			 	&48.15		&-		&-		&29.94 		\\																
SL  \cite{Chen2016Similarity}			 		&51.9		&-		&-		&26.35 		\\																
DNS  \cite{Zhang2016Learning}			 		&55.43		&-		&-		&29.87 		\\																
LSTM \cite{Varior2016A}			 	&61.6		&-		&-		&35.3 		\\																
S-CNN  \cite{Varior2016Gated}			 	&65.88		&-		&-		&39.55 		\\														
2Stream  \cite{Zheng2016A}			&79.51		&90.91		&94.09		&59.87 		\\																											
Pose  \cite{Zheng2017Pose}			 		&78.06		&90.76		&94.41		&56.23 		\\													
\hline																	
Baseline  			&78.2		&90.9		&94.4		&60.4 		\\
Our-I 			 	&79.3		&91.7		&94.9		&61.6		\\
Our-II 			 	&\textbf{79.6}		&\textbf{92.1}		&\textbf{95.0}		&\textbf{62.2} 		\\
\hline	
\end{tabular}
\end{table}

\begin{table}[htp!]
\center
\caption{ Person Re-ID Results on cuhk03 Benchmark.}\label{PersonREIDcuhk03benchmark}
\begin{tabular}{l|llll}
\hline
\hline
Methods  															&top-1  &top-5  &top-10 							\\
\hline
LMO \cite{Liao2015Person} 							&44.6				&- 					&-  				\\
DNS \cite{Zhang2016Learning} 				&62.6				&\textbf{90.0}				&\textbf{94.8}			\\
Gated-Siamese \cite{Varior2016Gated}							&61.8				&-					&-				\\
Siamese-LSTM \cite{Varior2016A}											&57.3				&80.1				&88.3			\\
Re-ranking\cite{Zhong2017Re}						&64.0				&-					&-				\\
\hline
Baseline  			&62.6			&87.7		&93.2		\\
Our-I 			 	&64.1   		&88.6 		&93.6 		\\
Our-II 			&\textbf{65.5}   		&89.3 		&94.0 		\\
\hline	
\end{tabular}
\end{table}

%% ------------------------------------------------------------------------------------
%%					Ablation Study 
%% ------------------------------------------------------------------------------------	
\subsection{Ablation Study}
To have a better understanding of our algorithm, we also implement ablation studies to validate the effectiveness of each component and related trade-off parameters. The detailed results can be found in following subsections. 

\textbf{The effectiveness of global mini-batch construction.} 
As shown in Table \ref{PersonREIDMarket1501} and Table \ref{PersonREIDcuhk03benchmark}, when we take the global mini-batch construction into consideration and improve the re-ID performance from rank-1 78.2\% to 79.3\% on Market1501 dataset, and from rank-1 62.6\% to 64.1\% on CUHK03 dataset (Our-I). It is a relative significant improvement when comparing with baseline method. This fully demonstrate the effectiveness of our global mini-batch construction.

\textbf{The effectiveness of improved triplet loss.} The baseline of our person re-identification is the results of original triplet loss function. As shown in Table \ref{PersonREIDMarket1501} and Table \ref{PersonREIDcuhk03benchmark}, this baseline achieves rank-1 78.2\% on Market1501 and rank-1 62.6\% on CUHK03 dataset, respectively. For the added regularization term, we can find that this term also improved the re-ID performance. On the basis of Our-I, when introduced the regularization term into the triplet loss function for feature learning, the final results can be improved from rank-1 79.3\% to 79.6\% on Market1501 benchmark and from 64.1\% to 64.7\% on CUHK03 dataset.
	
According to above observations, we can draw following conclusions: the global mini-batch construction based on human attribute recognition and the introduced regularization term are all improve the person re-ID performance on the two popular benchmarks.

\begin{figure*}[t]
\center
\includegraphics[width= 4.5in]{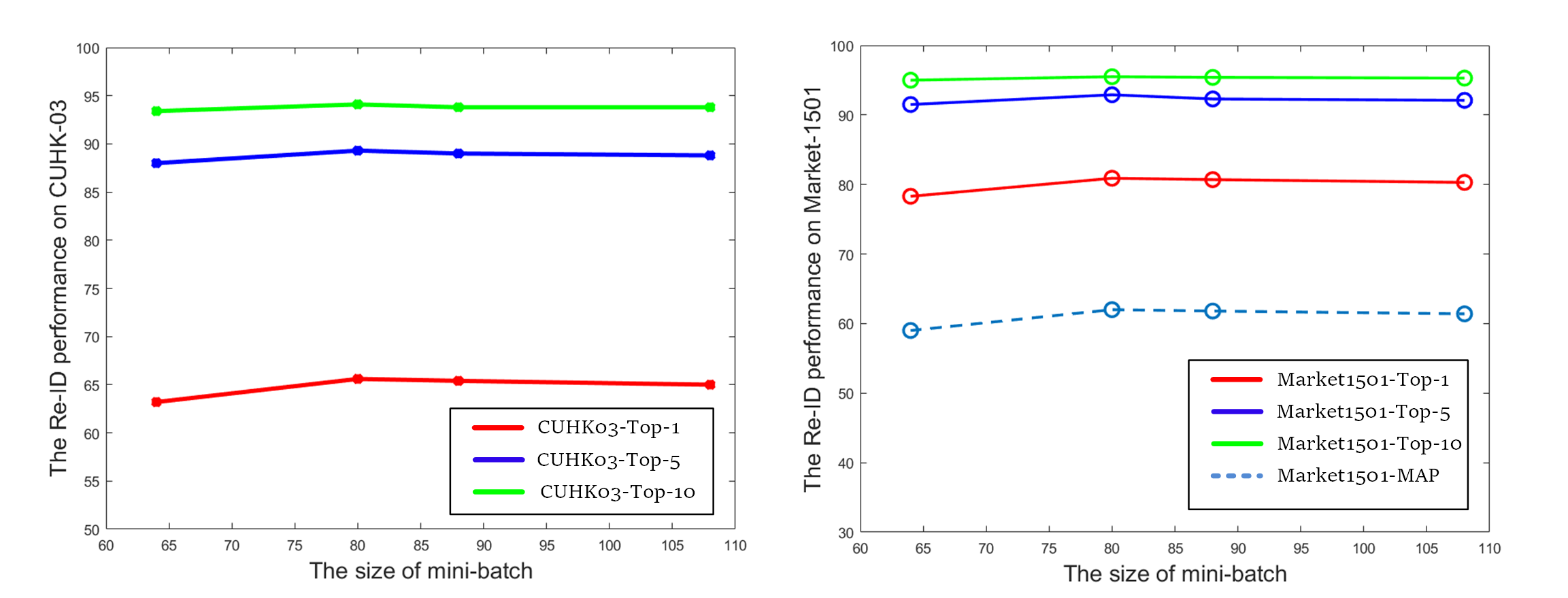}
\caption{The results with different mini-batch size on CUHK-03 and Market-1501 dataset.}
\label{ASMinibatchSizeCUHK}
\end{figure*}	
	
\textbf{The influence of mini-batch size.}
To validate the influence of different mini-batch size, we conduct the some experiments (\emph{i.e.} 64, 80, 88, 108), as illustrated in Figure \ref{ASMinibatchSizeCUHK}. It is easy to find that the final recognition performance can be enhanced when increasing the number of samples in each mini-batch. What's more, we can obtain better results when we set the batch size as 80.

\textbf{The results with different number of person IDs.}	
To check the influence of person IDs in one mini-batch, we also conduct experiments with different number of IDs. As shown in Table \ref{PersonREIDdifferpersonIDs}, we test our algorithm on Market-1501 dataset with $2, 4$ and $8$ instances and achieve 58.4, 62.2 and 63.3 on Mean AP, respectively. It is easy to find that our algorithm could achieve best result when the instance number is setting as 8.

\begin{table}[htp!]
\center
\caption{Results with different number of IDs on Market-1501 dataset (based on ResNet-50).}	
\label{PersonREIDdifferpersonIDs}
\begin{tabular}{c|ccccc}
\hline
\hline
Num-Instances  	&top-1  &top-5  &top-10 & Mean AP	\\
\hline 
8				&81.1 	&92.8	&95.5	&63.3		\\ 
4				&79.6	&92.1	&95.0	&62.2 		\\ 
2				&78.1 	&91.4	&94.3	&58.4		\\ 
\hline
\end{tabular}
\end{table}

\subsection{Limitations Analysis  and Future Works}\label{limitationsFutureworks}  
Although our proposed algorithm has achieved good performance on both person attributes recognition and pedestrian re-ID, however, our models still existing the following limitations. Firstly, we take the correlation between human attributes into considerations, but do not model the relations between the spatial location and specific attributes. This prior maybe have great improvement on human attribute recognition. Existing memory networks, such as recurrent neural network (RNN) or long short-term memory (LSTM) maybe a good choice to model these relations. Secondly, our method utilize the samples with high predicted score which selected by the proposed group driven multi-label self-paced learning algorithm in each iteration, and add these data to the training set. Our method still fail to get rid of the interference of wrong predictions in each iteration which is a common problem in semi-supervised learning. Actually, due to the existence of some tough samples which maybe predicted with some attributes with high confidence. However, these images may indeed not contain such attributes. This situation make the introduced noise in each iteration become unavoidable. More advanced algorithms maybe required  for selecting samples, such deep reinforcement learning \cite{Mnih2013Playing} \cite{Mnih2015Human}. We leave these two issues as our future works.

\section{Conclusion}\label{conclusions}
In this paper, we propose an improved hard example mining to further improve the training efficacy of models in person re-ID. We remark person identity and attribute recognition sharing a target at the human appearance description and propose an innovative hard person identity mining approach through a transferred human attribute encoder.  The encoder originated from a deep multi-task model robustly trained with a source noisy pedestrian attribute dataset, through a group prior driven self-paced learning in the semi-supervised manner. We apply this encoder to obtain each image attribute code in the target person re-ID dataset. Afterwards in the attribute code space, we consider each person as a latent identity distribution to generate his attribute codes as images in different practical scenarios. It presents as an complementary tool of hard example mining, which helps to explore the global instead of the local hard example constraint in the mini-batch built by randomly sampled identities. We validated the method on two person re-ID benchmarks, which both demonstrate the efficacy of our model-agnostic approach. We use PETA and extra unlabelled noisy data sources to attain our attribute encoder, which also outperforms various existing baselines in attribute recognition.

\section*{Acknowledgements}	
This work is jointly supported by National Natural Science Foundation of China (61702002, 61671018, 61872005), Key International Cooperation Projects of the National Foundation (61860206004), Natural Science Foundation of Anhui Province (1808085QF187), Natural Science Foundation of Anhui Higher Education Institution of China (KJ2017A017), Institute of Physical Science and Information Technology, Anhui University.

{
\bibliographystyle{plain} 
\bibliography{reference}
}

\end{document}